\pdfoutput=1

\documentclass[11pt]{article}
\usepackage[]{acl}
\usepackage{times}
\usepackage{latexsym}
\usepackage[T1]{fontenc}
\usepackage[utf8]{inputenc}
\usepackage{microtype}
\usepackage{inconsolata}
\usepackage{graphicx}
\usepackage{booktabs}
\usepackage{multirow}
\usepackage{amsmath}
\usepackage{algorithm}
\usepackage{algorithmic}
\usepackage{hyperref}
\usepackage{url}
\usepackage{enumitem}
\usepackage{xspace}
\usepackage{adjustbox}

\newcommand{\framework}{ICAT\xspace}

\newcommand{\factprecision}{S_\text{fact}}
\newcommand{\topicrecall}{S_\text{coverage}}
\newcommand{\factf}{\text{ICAT}}

\title{Beyond Factual Accuracy: Evaluating Coverage of Diverse Factual Information in Long-form Text Generation} 

\author{
  Chris Samarinas\textsuperscript{1}\thanks{Equal contribution.} \hspace{0.2cm}
  Alexander Krubner\textsuperscript{2}\footnotemark[1]\thanks{Work done while visiting CIIR at UMass Amherst.} 
  \hspace{0.2cm}
  Alireza Salemi\textsuperscript{1}\\
  \bf
  Youngwoo Kim\textsuperscript{1} \hspace{0.2cm}
  Hamed Zamani\textsuperscript{1} \\\\
  \textsuperscript{1}University of Massachusetts Amherst \quad \textsuperscript{2}Salzburg University of Applied Sciences \\
  \textsuperscript{1}\texttt{\{csamarinas, asalemi, youngwookim, zamani\}@cs.umass.edu} \\
  \textsuperscript{2}\texttt{akrubner.bin-m2022@fh-salzburg.ac.at} \\
}

\begin{document}
\maketitle

\begin{abstract}
This paper presents \framework,\footnote{source code: \url{https://github.com/algoprog/ICAT}} an evaluation framework for measuring coverage of diverse factual information in long-form text generation. \framework breaks down a long output text into a list of atomic claims and not only verifies each claim through retrieval from a (reliable) knowledge source, but also computes the alignment between the atomic factual claims and various aspects expected to be presented in the output. We study three implementations of the \framework framework, each with a different assumption on the availability of aspects and alignment method. By adopting data from the diversification task in the TREC Web Track and the ClueWeb corpus, we evaluate the \framework framework. We demonstrate strong correlation with human judgments and provide comprehensive evaluation across multiple state-of-the-art LLMs. Our framework further offers interpretable and fine-grained analysis of diversity and coverage. Its modular design allows for easy adaptation to different domains and datasets, making it a valuable tool for evaluating the qualitative aspects of long-form responses produced by LLMs.
\end{abstract}

\section{Introduction}
The rapid advancement of Large Language Models (LLMs) has revolutionized long-form text generation, enabling increasingly sophisticated applications from report writing to complex question answering. However, this progress has highlighted a critical challenge: how do we effectively evaluate not just the factual accuracy of generated content, but also its completeness and coverage of diverse perspectives? While recent work has made significant progress in developing comprehensive evaluation frameworks for LLMs \cite{llm-eval}, existing metrics often focus on isolated aspects like factuality or response quality \cite{min-etal-2023-factscore}, failing to capture the multi-dimensional nature of high-quality long-form text.

The evaluation of long-form text generation presents unique challenges that go beyond traditional metrics \cite{syntod}. Lexical overlap metrics like BLEU \cite{papineni2002bleu}, ROUGE \cite{lin2004rouge}, and METEOR \cite{banerjee2005meteor} are fundamentally limited by their reliance on surface-level text similarity, making them inadequate for evaluating semantically equivalent but lexically different expressions. While more recent approaches like BERTScore \cite{zhang2019bertscore} and G-Eval \cite{liu-etal-2023-g} attempt to address this through semantic similarity, they still face fundamental limitations when applied to long-form content. These metrics cannot effectively verify factual accuracy or assess whether the response comprehensively covers all relevant aspects of a topic. The vast space of possible acceptable outputs makes it impractical to create comprehensive reference texts, leading to the development of reference-free evaluation methodologies like Prism \cite{agrawal-etal-2021-assessing, thompson-post-2020-automatic}. However, these approaches often struggle to effectively identify hallucinations and biases.

Recent approaches like FActScore \cite{min-etal-2023-factscore} and VERISCORE \cite{song-etal-2024-veriscore} have addressed the factuality challenge by evaluating atomic claims against reliable sources, but factual accuracy alone is insufficient. A unified metric that considers both factuality and coverage is crucial not only for evaluation but also for optimizing LLM performance. Such a metric could serve as a reward function in reinforcement learning to simultaneously improve both the factual accuracy and comprehensive coverage of LLM outputs. 

Consider a user asking ``What are the health effects of coffee consumption?'' While an LLM response might present entirely factual claims about coffee's benefits, such as its role in improving alertness and potential protective effects against certain diseases, failing to address known health risks (like anxiety or sleep disruption) would present an incomplete and potentially misleading picture. This illustrates a critical gap in current evaluation approaches: the need to assess not just the accuracy of individual claims, but also the \emph{comprehensive coverage of diverse relevant information}. This is particularly crucial for applications like medical information systems, policy analysis, or educational content where balanced, complete information is essential for informed decision making.

This paper introduces \framework, a novel evaluation framework that addresses this gap by measuring both factual accuracy and coverage of diverse factual information in long-form text generation. Our key contributions include: 1) a modular evaluation framework that decomposes long-form text into atomic claims and evaluates both their factual accuracy and their alignment with expected aspects, 2) three implementations with varying degrees of automation, suitable for different evaluation scenarios, and 3) a comprehensive evaluation of various LLMs and demonstrating strong correlation with human judgments in a user study.

\framework first breaks down the generated long text into atomic claims. Through retrieval from a (reliable) corpus $C$ or the Web, \framework verifies each atomic claim to ensure its factuality. To measure completeness and coverage of diverse facts, \framework requires \textit{a set of diverse aspects} to compute an \textit{alignment} between each atomic factual claim in the LLM response and the set of diverse aspects. We study three implementations of the \framework framework with different assumptions on the availability of aspects and alignment methods. 

In our experiments, we use ClueWeb \cite{clueweb09} as the retrieval corpus. We solely focus on the English documents of the ClueWeb collection. In addition to custom open-source retrieval from ClueWeb, we explore web-based grounding using the Brave Search API. For experiments, we rely on the input queries from the TREC Web Track \cite{trec09, trec10, trec11, trec12}. The argument for this decision is based on the fact that TREC Web Track queries have also been used for search result diversification. This means that the queries include up to 7 aspects and documents are provided with aspect-level relevance annotations. Our experiments show that there is relatively strong corelation of ICAT with human judgments, showcasing the utility of this framework for evaluating coverage of diverse factual information in LLM responses without human input.

By offering a modular and adaptable framework, \framework enables researchers to tailor the evaluation process to specific needs, making it a valuable tool for assessing the qualitative aspects of long-form responses produced by LLMs. The decomposition of LLM outputs into atomic claims and their alignment with specific topics makes the evaluation process highly interpretable - evaluators can trace exactly which claims support which topics and identify gaps in coverage. This granular analysis capability, combined with the framework's ability to evaluate both factual accuracy and topic coverage, provides a more comprehensive assessment compared to existing metrics that only measure one of these aspects. 

\begin{figure*}
    \centering
    \includegraphics[width=0.82\textwidth]{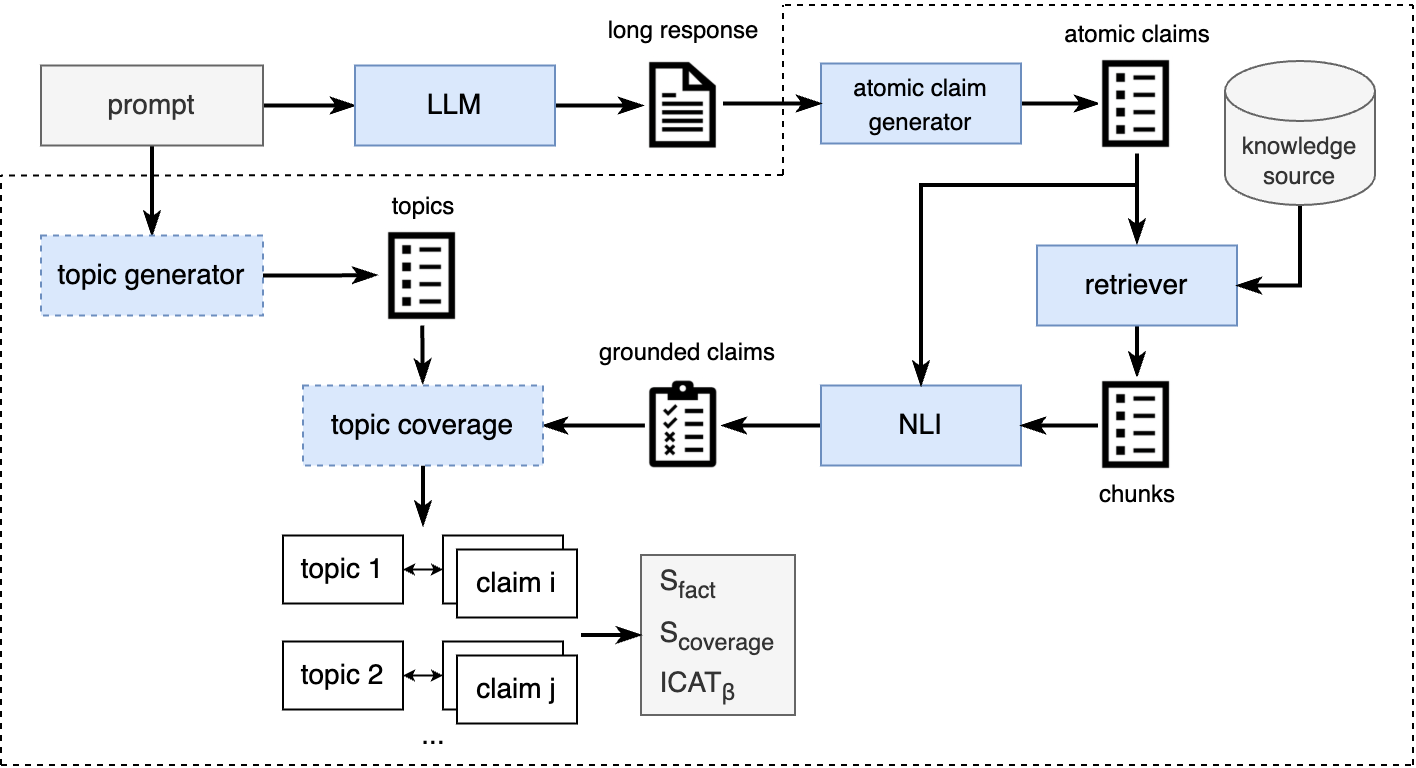}
    \caption{Retrieval-based evaluation of LLM responses with \framework. Topic generation and coverage models are optional depending on the chosen evaluation method.}
    \label{fig:coverage-score-a}
\end{figure*}

\section{Related Work}
\textbf{Text Generation Evaluation} Traditional approaches to evaluating generated text have primarily focused on n-gram overlap metrics such as BLEU \cite{papineni2002bleu}, ROUGE \cite{lin2004rouge}, and METEOR \cite{banerjee2005meteor}. While these metrics are effective for assessing local coherence and fluency, they fail to capture higher-level aspects such as topic coverage and diversity. Recent work has introduced more sophisticated metrics like BERTScore \cite{zhang2019bertscore}, BLEURT \cite{sellam2020bleurt}, and unified multi-dimensional evaluators \cite{zhong-etal-2022-towards} which leverage pre-trained LLMs for more nuanced evaluation.

\textbf{Topic Coverage and Diversity} Research on evaluating topic coverage has roots in information retrieval, where metrics like $\alpha$-nDCG \cite{clarke2008novelty} and S-Recall were used to assess the topical diversity of search results. The concept of diversity in evaluating generated text can encompass various interpretations, including lexical diversity (analyzing the variety of words used) and topical diversity (assessing the range of topics covered). In the context of text generation, recent work has explored various approaches to measuring lexical diversity, including term overlap self-similarity such as Self-BLEU \cite{self-bleu} and the proportion of distinct unigrams and bigrams in generated responses \cite{li-etal-2016-diversity}. However, research on evaluation of topical diversity in LLMs is currently limited.

\textbf{LLM Evaluation Frameworks} Several frameworks have been proposed for evaluating different aspects of LLM performance, including factuality \cite{min-etal-2023-factscore,song-etal-2024-veriscore} and dialogue quality \cite{mehri2020usr}. Concurrent to this research, the AutoNuggetizer framework \cite{pradeep2024initialnuggetevaluationresults} used LLMs to generate and assess the coverage of nuggets in text. The EXAM++ framework \cite{farzi2024exambasedevaluationapproachtraditional} evaluates information coverage by checking if system responses can answer query-related exam questions, focusing on high-level answerability. In contrast, our work provides more fine-grained explainability by aligning individual atomic claims with retrieved evidence to verify both factual accuracy and aspect coverage at a more granular level. This allows us to not only assess if an aspect is covered, but also identify the specific claims and evidence supporting that coverage.

Our work builds upon these foundations, specifically addressing the challenge of evaluating topic coverage in long-form text generation while considering factuality at the same time.

\section{\framework}
\label{sec:method}

\begin{table*}[t]
    \centering
    \resizebox{\textwidth}{!}{
    \begin{tabular}{lll}\toprule
         &  \textbf{Approach for Obtaining Diverse Aspects} & \textbf{Approach for Claim-Aspect Alignment} \\\midrule
        \textbf{\framework-M} & Manual: ground-truth aspects & Manual: retrieval-based method with aspect-level ground-truth alignment \\
        \textbf{\framework-S} & Manual: ground-truth aspects & Automatic: retrieval-based method with aspect-level LLM-based alignment \\
        \textbf{\framework-A} & Automatic: LLM-based aspect generation & Automatic: retrieval-based method with aspect-level LLM-based alignment \\\bottomrule
    \end{tabular}
    }
    \caption{The methods used for obtaining query aspects and claim-aspect alignment in each variant of \framework.}
    \label{tab:icat_variants}
\end{table*}

Queries that require a long-form response, e.g., complex non-factoid questions, are often associated with multiple aspects. The response to these queries often include multiple claims, some of which  may be factually accurate, while others may be inaccurate. An ideal response to these queries should not only contain factually accurate claims, but should also leave no aspect or perspective unaddressed. For instance, an ideal answer to a question about a legislation should cover perspectives from all political parties. An ideal answer to a question about the impact of a food or a medication on health should cover both positive, neutral, and negative perspectives. However, no existing evaluation metric can evaluate both factual accuracy and aspect coverage in long-form text generation. To address these, given a long output $y$ produced in response to an input $x$, the \framework framework computes two main scores: \textit{factuality score} and \textit{coverage score}.

\paragraph{Factuality Score.} Building upon prior work, such as FActScore \cite{min-etal-2023-factscore} and VERISCORE \cite{song-etal-2024-veriscore}, Factuality Score measures the ratio (or percentage) of factually accurate claims in $y$. 
To do so, it is crucial that the generated claims are accurate. Let $AC(y)$ be a function that extracts atomic claims from a generated response $y$. Given the set of atomic claims $C = AC(y)$ made in $y$, we define the function $C_T = CG(C; K)$ that verifies the factuality of claims in $C$ using a given knowledge source $K$. Therefore, $C_T \subseteq C$ denotes the set of factually verified claims in response $y$. Factuality Score is then defined as follows:
\[
\factprecision = \frac{|C_T|}{|C|}
\]
where $|\cdot|$ denotes the cardinality of the given set.

\paragraph{Coverage Score.}
To evaluate information coverage and diversity in $y$, Coverage Score measures the ratio of aspects being covered by the factually accurate claims in the given text.
Hence, it is essential to identify which query aspects are accurately addressed in the generated response. Formally, coverage score can be defined as:
\[
\topicrecall = \frac{|\{TO(c, K):  c \in C_T \} \cap TQ(x)|}{|TQ(x)|}
\]
where $TO$ is a function that identifies the subtopics associated with claim $c$, and $TQ$ is a function that returns all aspects related to the input $x$. Note that aspect coverage is only computed for factually verified claims, i.e., $C_T$, instead of all claims. The reason is that non-factual claims should be avoided, regardless of the aspect they cover. Thus, they should not contribute to the coverage score. 

\paragraph{The $\factf_\beta$ Score.}
Inspired by F-measure \cite{van1979information}, we calculate the weighted harmonic average of these factuality and coverage scores, as follows:
\[
\factf_\beta = (1+\beta^2)\frac{\factprecision \cdot \topicrecall}{\beta^2 \factprecision + \topicrecall}
\]
where parameter $\beta$ is a hyper-parameter that controls the trade-off between the factuality and coverage scores. In more detail, $\beta$ controls the weight of Coverage Score compared to Factuality Score. Thus, a higher $\beta$ signifies the impact of information coverage, while a lower $\beta$ prioritizes factual accuracy. The default value for $\beta$ is equal to $1$, where factuality and coverage score are weighted uniformly. Throughout this paper, when the value of $\beta$ is not explicitly mentioned, the default value of $1$ is being used. 

\paragraph{Variants of \framework}
We study three variants of \framework implementations based on how they obtain query aspects and compute the alignment between atomic claims and aspects. Table~\ref{tab:icat_variants} describes the approaches used in these three variants and highlights their differences. \framework-M assumes that a ground-truth set of diverse aspects are obtained manually and is available to the evaluation framework. It also assumes that ground-truth relevance annotations for each document in the corpus $C$ to each aspect are provided. Using this information, the retrieval model can identify which aspect is being covered by each atomic factual claim in the LLM response. \framework-S similarly assumes that a ground-truth set of diverse aspects are obtained manually, however no aspect-level relevance judgment is available. Therefore, it uses an LLM to conduct pseudo-labeling and perform alignment between the atomic factual claims and the set of aspects. \framework-A assumes that the aspect set is not available, so it first uses an LLM to automatically generate diverse aspects of the input and then conducts pseudo-labeling for alignment.

The rest of this section provides details on how to develop models for generating factual claims (i.e., function $AC$), how to validate the factuality of claims (i.e., function $CG$), and how to obtain query aspects and compute an alignment between factual claims and all query aspects (i.e., functions $TQ$ and $TO$). The rest of this section describes the approaches at high level to introduce the generic \framework, while Section~\ref{sec:details} provides the implementation details used in our experiments.

\subsection{Atomic Claim Generation}
\begin{figure}
    \centering
\includegraphics[width=0.485\textwidth]{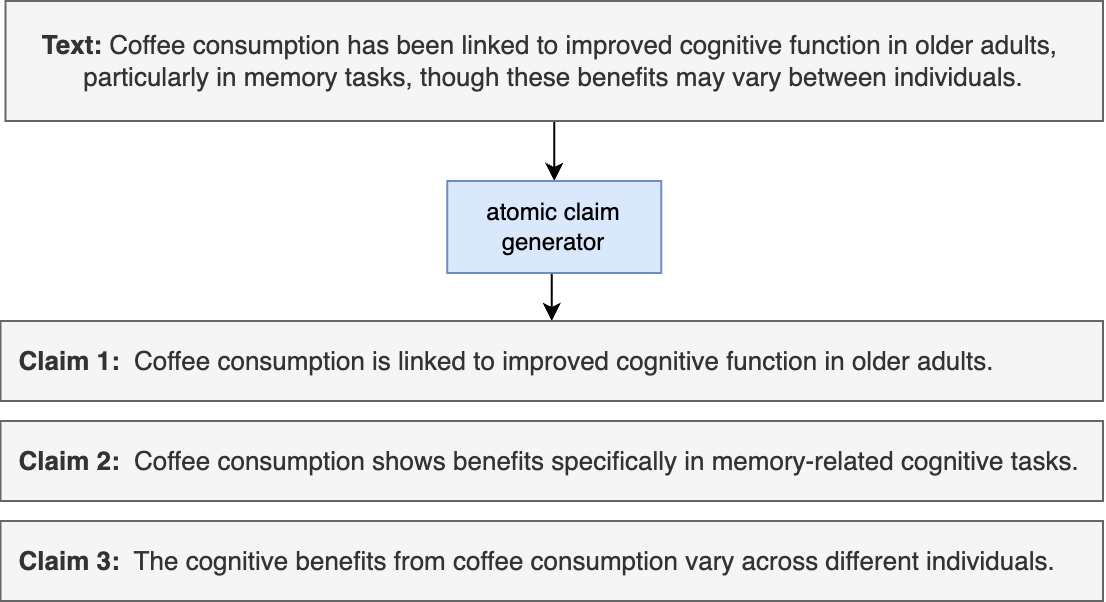}
    \caption{Example of atomic claim generation}
    \label{fig:claim-generator}
\end{figure}

The atomic claim generation process seeks to break down a given long text into standalone and atomic claim statements that preserve key context and maintain claim consistency (see Figure~\ref{fig:claim-generator}). The generated claims should strike an appropriate balance of granularity, ensuring they are self-contained and decontextualized. We assume the existence of a function \( C = AC(y) \), which returns a set of atomic claims \( C \), given the long output text $y$. There are various ways to implement this; one might consider each sentence or paragraph in $y$ as an atomic claim. However, this simple approach does not satisfy our expected self-containment and decontextualization qualities. Instead, we follow \cite{min-etal-2023-factscore} and utilize an LLM \( M_{\text{claims}} \) with the prompt shown in Figure~\ref{fig:atomic-claim-prompt}. This prompt instructs the LLM to decompose the generated response \( y \) into multiple self-explanatory and decontextualized sentences, each containing a single atomic fact. These sentences then constitute the set of atomic claims for the generated output, denoted as $C = AC(y)$. An example of this process is illustrated in Figure~\ref{fig:claim-generator}. An instruction-tuned LLM can be used as \( M_{\text{claims}} \) in a zero- of few-shot setting; however, we found that smaller-scale LLMs (such as LLaMA \cite{llama} with 8 billion parameters) cannot accurately perform this task without fine-tuning. Therefore, we either use an LLM with higher capacity or distill knowledge into a smaller-scale LLM, enabling faster inference for our framework. The details of this distillation process are provided in section~\ref{sec:claim-gen}.

\subsection{Claim Grounding}
\label{sec:claim-grounding}

To design the claim grounding function \( CG \), for each claim \( c \in C \), we employ a retrieval model \( R \) to retrieve \( n \) documents from the given knowledge source \( K \). Subsequently, a natural language inference (NLI) model \( M_{\text{NLI}} \) is used to determine whether the claim can be supported by any of the retrieved documents. If the claim can be inferred from at least one of the retrieved documents, it is considered grounded (i.e., validated, thus factually accurate); otherwise, it is not. The function returns a subset of $C$ that are found grounded. 

\subsection{Aspect Coverage Assessment}

To calculate aspect coverage for some given input prompt (query) \( x \), it is essential to have a list of diverse aspects for \( x \) (i.e., $TQ(x)$) and a method to determine which aspect each claim pertains to (i.e., $TO(c, K): c\in C$). 

\paragraph{Methods for Obtaining Diverse Query Aspects ($TQ$):} 

We propose two main methods to identify all aspects related to the query \( x \):  
\begin{itemize}[leftmargin=1em]
    \item \textbf{Manual--Ground-truth Aspects:} In this case, the aspects that should be included in the response to the query \( x \) are provided as a reference for evaluation. 
    \item \textbf{Automatic--LLM-based Aspect Generation:} Building on previous work showing LLMs can effectively identify aspects of a query \cite{10.1145/3539813.3545138}, we use an LLM \( M_{\text{subtopic}} \) with the prompt shown in Figure~\ref{fig:topic-generation-prompt}. This prompt instructs the LLM to generate up to 10 aspects for the query, covering the key aspects about it. This approach is useful when ground-truth aspects are unavailable.
\end{itemize}

\paragraph{Methods of Obtaining the Aspects of an Atomic Claim ($TO$):}

We use two methods to identify the aspects related to an atomic claim $c$:
\begin{itemize}[leftmargin=1em]
    \item \textbf{Manual--retrieval-based method with aspect-level ground-truth alignment:} In this method, we assume access to a knowledge source \( K \), where each document is annotated with the aspects it covers. To find the aspects that the claim \( c \) covers, we use the retrieval model \( R \) to retrieve \( n \) documents. Then, according to the ranking, we find the first ranked document that supports the claim \( c \) using the method in Section \ref{sec:claim-grounding}. The aspects of this document are considered as the aspects that the claim \( c \) covers. If none of the documents support claim \( c \), we assume that it does not cover any query aspect. 
    
    \item \textbf{Automatic--retrieval-based method with aspect-level LLM-based alignment:} In this method, we use an aspect-claim alignment LLM $M_{\text{coverage}}$ to determine which aspects each grounded claim covers. Given a query $x$, its aspects $TQ(x)$, and a set of grounded claims $C_T$, we prompt the LLM to analyze each claim and identify which aspects it addresses. The prompt (shown in Figure~\ref{fig:coverage-prompt}) instructs the LLM to output a structured mapping between claims and aspects, where each claim can be mapped to zero, one, or multiple aspects. 
    This approach eliminates the need for aspect-level relevance judgments in the knowledge source while still maintaining a retrieval-based verification of factual accuracy. Unlike the manual method that assumes a claim covers the aspects associated with its supporting document, this method directly analyzes the semantic relationship between claims and aspects, leading to more accurate assessment.
    
\end{itemize}

\begin{table*}[ht!]
\centering
\scriptsize
\begin{tabular}{ll|ccc|ccc}
\toprule
& & \multicolumn{3}{c|}{\textbf{Corpus-based Retrieval}} & \multicolumn{3}{c}{\textbf{Web-based Retrieval}} \\
\midrule
\textbf{Method} & \textbf{Coverage Model} & \textbf{Pearson's $\rho$} & \textbf{Spearman's $\rho$} & \textbf{Kendall's $\tau$} & \textbf{Pearson's $\rho$} & \textbf{Spearman's $\rho$} & \textbf{Kendall's $\tau$} \\\midrule
\framework-M & N/A & 0.196 & 0.184 & 0.146 & - & - & - \\
\midrule
\multirow{2}{*}{\framework-S} & Llama-3.1-8B & 0.247 & 0.255 & 0.200 & 0.294 & 0.252 & 0.204 \\
& Llama-3.1-70B & \textbf{0.422} & \textbf{0.446} & \textbf{0.368} & \textbf{0.525} & \textbf{0.443} & \textbf{0.358} \\
\midrule
\multirow{2}{*}{\framework-A} & Llama-3.1-8B & 0.161 & 0.125 & 0.097 & 0.033 & 0.021 & 0.014 \\
& Llama-3.1-70B & 0.246 & 0.218 & 0.173 & 0.055 & 0.054 & 0.045 \\
\bottomrule
\end{tabular}
\caption{Correlation of information coverage in the proposed evaluation methods with manual human annotations. Though ICAT-A shows weaker correlation than ICAT-S, we observed that auto-generated topics were more comprehensive and higher quality than the existing ones in the TREC dataset, suggesting ICAT-A would perform better with more complete ground-truth topics.}
\label{tab:correlations}
\end{table*}

\section{Implementation Details}
\label{sec:details}

\subsection{Atomic Claim Generation}
\label{sec:claim-gen}
The claim generation module was trained with several key objectives in mind, building on recent work in atomic claim extraction \cite{min-etal-2023-factscore}. The model learned to extract standalone factual statements from text while maintaining factual consistency and simplifying complex statements. Special attention was paid to preserving important context and qualifiers, and generating claims at an appropriate granularity level \cite{song-etal-2024-veriscore}. For this task we used Llama 3.1 8B \cite{llama} fine-tuned using QLoRA \cite{dettmers2023qlora} on synthetic examples. We tried using models of this size without fine-tuning, however we found that the generated claims are often not de-contextualized properly. Larger models with 70B or more parameters seem to be effective for this task without fine-tuning, however they are very expensive to run, especially for long texts.

The synthetic training data was generated through a multi-stage process by prompting Llama 3.1 405B. We began by generating 200 diverse high-level topics across multiple domains. For each topic, we generated 5 relevant entities. We then created variable-length paragraphs for each entity and generated the associated list of atomic claims for each of them. Using these 1000 synthetic examples, we fine-tuned the model for 1 epoch with batch size 16, learning rate 2e-4 and LoRA parameters $\alpha$ = 16 and rank = 64.

Our human evaluation comparing the fine-tuned 8B model with zero-shot 70B model revealed an interesting trade-off: while the 70B model achieves better semantic similarity with ground-truth facts, the fine-tuned 8B model produces better-formulated claims with higher fact precision (0.838 vs 0.753). This superior decontextualization is crucial for accurate retrieval-based grounding, as claims must be self-contained to match against corpus snippets effectively. This finding, combined with the significant computational cost savings (approximately 8.75× less memory and proportionally faster inference), justifies our design choice of using a fine-tuned smaller model. We plan to release a similarly fine-tuned adapter for fact-topic alignment to close the performance gap with the 70B model in that component as well. A detailed evaluation of the atomic claim generation quality is provided in Appendix~\ref{app:component-eval}.

\subsection{Topic Generation}

Previous works have shown that LLMs can be very effective in query subtopic generation \cite{10.1145/3539813.3545138}. In our framework, for generating ground-truth topics given a query, we use the same base LLM as the one used in claim generation. Here we found that even without fine-tuning, Llama 3.1 8B can produce relevant topics. In order to reduce the need for extra resources to use a base and fine-tuned version of the LLM for claim generation, we use the VLLM library \cite{kwon2023efficient} to load the base model only once in memory and efficiently serve the adapter for the fine-tuned version. The quality evaluation of topic generation is detailed in Appendix~\ref{app:component-eval}.

\subsection{Claim Grounding}

We implemented a two-stage approach for grounding atomic claims in the given text with a corpus. We first preprocess the corpus and generate chunks for each document with up to 128 words with 32 words overlap. We use a dense embedding model\footnote{\url{hf.co/Snowflake/snowflake-arctic-embed-m}} \cite{Merrick2024ArcticEmbedSE} to produce embeddings for all snippets and FAISS \cite{faiss} to build an efficient approximate nearest-neighbor index. We used IVF with HNSW for cluster assignment as our index type for fast search even when providing a large-scale corpus.

In the first stage, a retriever is used to obtain the $k=10$ most relevant snippets in the corpus for each claim. When web search is used instead of a corpus, we use the returned snippets from Brave Search API. In the second stage, a natural language inference model is used to filter only the supported claims. We use a model based on DeBERTa V3 \cite{he2021debertav3} fine-tuned on MultiNLI, FEVER and Adversarial NLI \cite{mnli, fever, anli}. A claim is kept if there is at least one snippet that supports it. Instead of using a LLM for filtering the supported claims, we use a much smaller BERT based model \cite{devlin-etal-2019-bert} fine-tuned specifically for this task.

We limit the snippet length because both NLI and dense embeddings models based on small pre-trained transformer LMs like BERT tend to have lower performance as the input length increases. The choice of using a specialized NLI model rather than a zero-shot LLM for claim verification was driven by both efficiency and effectiveness considerations. Our DeBERTa-based model achieves strong performance across multiple NLI benchmarks (MNLI: 0.903, FEVER-NLI: 0.777, ANLI-all: 0.579) while being orders of magnitude faster and more resource-efficient than LLM-based approaches. Given that grounding requires processing numerous claim-snippet pairs, efficiency is crucial for practical deployment and potential use as a reward function in reinforcement learning optimization of LLMs. A comprehensive evaluation of the NLI model performance across benchmarks is provided in Appendix~\ref{app:component-eval}.

\subsection{Aspect-Claim Alignment}
Topic coverage is assessed using the same base LLM with claim and topic generation. Given a query, a list of enumerated atomic claims and a list of ground truth topics, the LLM is prompted to produce a list of covered topic ids with their associated claim ids in structured jsonl format. The accuracy of the aspect-claim alignment module is evaluated in Appendix~\ref{app:component-eval}.

\begin{table*}[ht!]
    \scriptsize
    \centering
    \adjustbox{max width=\textwidth}{
    \begin{tabular}{l|c|cc|cc|cc}
    \toprule
    & & \multicolumn{2}{c|}{\textbf{ICAT-M}} & \multicolumn{2}{c|}{\textbf{ICAT-S}} & \multicolumn{2}{c}{\textbf{ICAT-A}} \\
    \cmidrule(lr){3-4} \cmidrule(lr){5-6} \cmidrule(lr){7-8}
    \textbf{LLM} & $\factprecision$ & $\topicrecall$ & $\factf\text{-M}_1$ & $\topicrecall$ & $\factf\text{-S}_1$ & $\topicrecall$ & $\factf\text{-A}_1$ \\
    \midrule
    \multicolumn{8}{c}{{Corpus-based Retrieval}} \\
    \midrule
    GPT-4 & 0.343 & 0.416 & 0.327 & 0.453 & 0.346 & 0.463 & 0.354 \\
    Llama-3-70B & 0.327 & 0.451 & 0.335 & 0.464 & 0.355 & 0.466 & 0.354 \\
    Mixtral-8x22B & 0.344 & 0.370 & 0.297 & 0.414 & 0.342 & 0.409 & 0.339 \\
    Openchat 3.5 (7B) & 0.340 & 0.413 & 0.329 & 0.429 & 0.348 & 0.424 & 0.347 \\
    \midrule
    \multicolumn{8}{c}{{Web-based Retrieval}} \\
    \midrule
    GPT-4 & 0.748 & - & - & 0.551 & 0.634 & 0.541 & 0.627 \\
    Llama-3-70B & 0.714 & - & - & 0.556 & 0.625 & 0.486 & 0.578 \\
    Mixtral-8x22B & 0.749 & - & - & 0.503 & 0.601 & 0.429 & 0.545 \\
    Openchat 3.5 (7B) & 0.741 & - & - & 0.520 & 0.611 & 0.444 & 0.555 \\
    \bottomrule
    \end{tabular}}
    \caption{Evaluation of various LLMs using ICAT. Llama-3.1-70B is used for claim-aspect alignment.}
    \label{tab:model-performance-all}
\end{table*}

\section{Experimental Setup}

\paragraph{Dataset.} We conducted our experiments using the ClueWeb09 Category B corpus--a large-scale web collection with over 50 million English documents \cite{clueweb09}. This corpus has been used in TREC Web Track from 2009 to 2012 \cite{trec09, trec10, trec11, trec12}, consisting of 200 topics, derived from a commercial search engine's query log, balanced for popularity. Each topic includes a topic title (i.e., often used as the keyword search query), a description (i.e., detailed description of the information need), type, and subtopics (i.e., diverse aspects of the topic). The relevance judgments encompass 38,637 query-document pairs, with 19.06\% (7366) marked as relevant. The dataset's unique advantage lies in its comprehensive coverage of internet content and human-annotated relevance judgments for topical diversity assessment. Relevance was judged either binary or on a five-point scale (later converted to binary), with documents considered relevant when containing useful information for specific subtopics. In our experiments with this collection, we filtered out spam documents using the Waterloo spam scorer \cite{spam} with the threshold of 70\%. We used BM25 to retrieve 1000 documents for each topic (given its title as the query string) and considered these documents for retrieval in our factual verification process. From the original 200 TREC topics, we filtered out approximately 21 queries that were not suitable for long-form text generation evaluation (e.g., queries asking for images, URLs, or specific data points), resulting in 179 valid queries. For human evaluation and detailed component analysis, we randomly sampled 50 queries from this filtered set due to cost constraints. These 50 queries were also used for LLM evaluation to ensure consistency and enable direct comparison with human judgments.

\paragraph{Experimental Setup.} We evaluated four state-of-the-art LLMs: GPT-4, Llama-3-70B-Instruct, Mixtral-8x22B-Instruct-v0.1, and Openchat 3.5 (a fine-tune of Mistral-7B) \cite{wang2023openchat}. For each model, we generated responses for each test query. For the baselines in this paper, we used the query descriptions in their original format from the ClueWeb09 dataset as prompts, which are not optimized for producing very diverse outputs. 





\section{Experimental Results}

\paragraph{Human Evaluation Study.}
To validate \framework's effectiveness, we conducted a comprehensive human evaluation study using Amazon Mechanical Turk (AMT). For each query-answer pair, three independent annotators assessed the coverage of aspects through a custom interface (Figure~\ref{fig:annotation-ui} in Appendix~\ref{app:human-annotation}). We limited the HITs to adult workers from the US, UK, Australia and Ireland, with over 98\% approval rate who have completed at least 5,000 assignments. The annotators were tasked with identifying whether specific aspects are present in a given LLM-generated text and highlighting corresponding text evidence for each identified aspect. To ensure quality annotations, we provided detailed guidelines with two reference examples. We use majority voting across annotators. The study achieved substantial inter-annotator agreement with Fleiss's $\kappa$ = 0.829, which is considered as a substantial agreement. For each query, we calculated Coverage Scores based on the set of covered topics identified by each evaluation method (ICAT variants) and by human annotators, relative to the set of ground truth topics. These per-query coverage scores were then used to compute linear and rank-bsaed correlation metrics (i.e., Pearson's $\rho$, Spearman's $\rho$, and Kendall's $\tau$) between the automated ICAT methods and human judgments. 

The correlation analysis between \framework variants and human judgments (see Table~\ref{tab:correlations}) reveals strong performance across most evaluation methods. Using Llama-3.1-70B as the coverage model, \framework-S achieves the strongest correlations (Pearson's $\rho$ = 0.422, p < 0.01; Spearman's $\rho$ = 0.446, p < 0.01). Interestingly, \framework-A demonstrates better performance than ICAT-M despite not requiring ground-truth aspects. While ICAT-A seems to have much weaker correlation than ICAT-S, it is worth noting that in practice, we observed the automatically generated topics to be higher quality and more exhaustive. Using a more comprehensive set of ground-truth topics, ICAT-A would probably demonstrate higher correlation with human judgements.

\paragraph{Comparison with Traditional Metrics for Text Generation.}
To demonstrate the advantages of \framework over traditional text generation evaluation metrics for assessing topic coverage, we compared our approach with widely-used metrics including BLEU, ROUGE, METEOR, and BERTScore. For this comparison, we concatenated the text of all ground-truth topics as a reference and evaluated the LLM responses against this reference. Table~\ref{tab:traditional-metrics} presents the correlation results with human-annotated coverage scores. 

Among traditional metrics, only BERTScore-recall shows statistically significant correlation with ground-truth coverage scores (Pearson's $\rho$ = 0.291, p < 0.001). However, this correlation is still much lower than ICAT-S. Notably, n-gram based metrics (BLEU, ROUGE variants) show negative or near-zero correlations, demonstrating their inadequacy for evaluating topic coverage in long-form text. These results confirm that traditional metrics, designed primarily for surface-level text similarity, fail to capture the semantic coverage of diverse aspects that \framework effectively measures.

\begin{table*}[ht!]
\centering
\scriptsize
\begin{tabular}{l|cc|cc|cc}
\toprule
\textbf{Metric} & \textbf{Pearson $\rho$} & \textbf{p-value} & \textbf{Spearman $\rho$} & \textbf{p-value} & \textbf{Kendall $\tau$} & \textbf{p-value} \\
\midrule
BERTScore-precision & -0.003 & 0.966 & 0.048 & 0.499 & 0.019 & 0.709 \\
BERTScore-recall & 0.291 & 0.000 & 0.313 & 0.000 & 0.226 & 0.000 \\
BERTScore-F1 & 0.120 & 0.090 & 0.188 & 0.008 & 0.126 & 0.013 \\
METEOR & 0.027 & 0.704 & 0.100 & 0.160 & 0.071 & 0.164 \\
ROUGE-1 & -0.058 & 0.415 & 0.016 & 0.819 & 0.014 & 0.785 \\
ROUGE-2 & -0.090 & 0.204 & -0.032 & 0.655 & -0.018 & 0.725 \\
ROUGE-L & -0.090 & 0.206 & 0.020 & 0.781 & 0.012 & 0.808 \\
ROUGE-Lsum & -0.090 & 0.206 & 0.020 & 0.781 & 0.012 & 0.808 \\
BLEU & -0.099 & 0.165 & -0.059 & 0.409 & -0.046 & 0.413 \\
\bottomrule
\end{tabular}
\caption{Correlation of traditional text generation metrics with human-annotated topic coverage scores. Only BERTScore-recall shows statistically significant correlation, though still lower than \framework variants.}
\label{tab:traditional-metrics}
\end{table*}

\paragraph{Comparing factuality and coverage of information in state-of-the-art LLMs using \framework.}
Our experimental results reveal distinct patterns in how different LLMs balance factuality and coverage (see Table~\ref{tab:model-performance-all}). When using corpus-based retrieval, Llama-3-70B demonstrates superior Coverage Score (0.451), while GPT-4 and Mixtral-8x22B show comparable factuality scores (0.343 and 0.344, respectively). However, Mixtral exhibits notably lower Coverage Score (0.370) compared to GPT-4's 0.416, resulting in lower overall $\factf_1$ scores (0.297 vs 0.327). Notably, when employing web-based retrieval, we observe substantially higher factuality scores across all models. This significant improvement can be attributed to the broader knowledge base available through web search, allowing more claims to be successfully grounded. Coverage scores also show improvement with web-based retrieval, though the increase is more modest. This suggests that while web search enables better fact verification, the comprehensiveness of aspect coverage is more dependent on the model's capabilities than the retrieval source.

\begin{figure}
    \centering
    \includegraphics[width=0.45\textwidth]{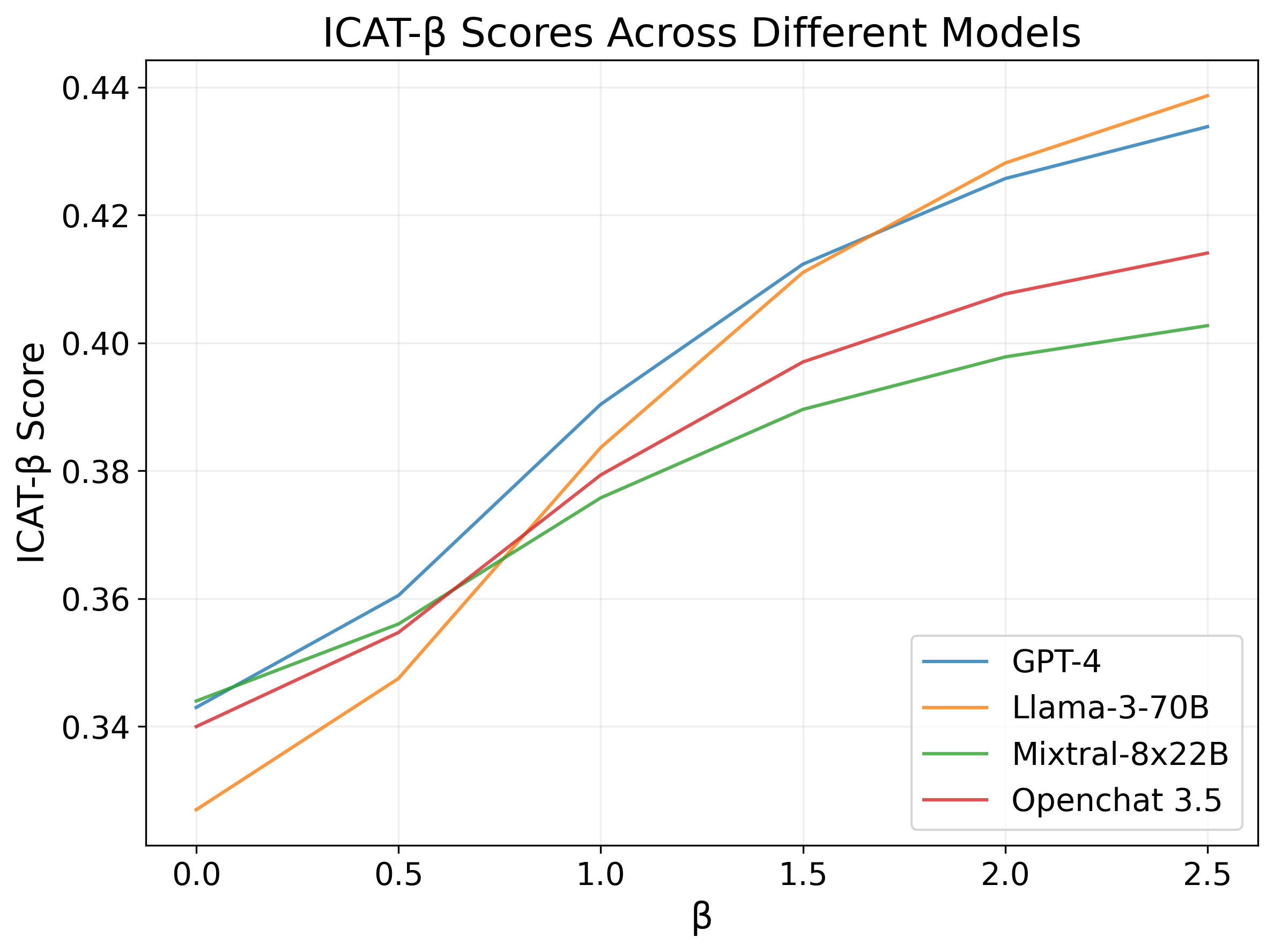}
    \caption{\text{\framework-A}$_{\beta}$ for various LLMs using Llama-3.1-70B as coverage model and the corpus as knowledge source. For low $\beta$ values where factuality has higher weight, GPT-4 and Mixtral have better performance while for higher values of $\beta$, GPT-4 and Llama have higher scores due to higher $S_{\text{coverage}}$. 
    }
    \label{fig:coverage-score-b}
\end{figure}

\paragraph{Controlling the trade-off between factuality and coverage using $\beta$.}
Figure~\ref{fig:coverage-score-b} illustrates how different values of $\beta$ affect the trade-off between factuality and coverage scores. Users can adjust $\beta$ based on their evaluation priorities: values of $\beta < 1$ give more weight to factuality score, while $\beta > 1$ emphasizes on coverage. 
In our experiments, for lower $\beta$ values, GPT-4 and Mixtral demonstrate superior performance in terms of $\factf_\beta$. However, as $\beta$ increases and coverage becomes more important, GPT-4 and Llama achieve higher scores due to their stronger Coverage Score.

\paragraph{Component-level Evaluation.}
To understand the contribution of each component in our framework, we conducted detailed evaluations of the atomic claim generation, topic generation, and coverage assessment modules. A comprehensive analysis of each component's performance is provided in Appendix~\ref{app:component-eval}.

\section{Conclusions and Future Work}

We presented \framework, a comprehensive framework for evaluating topic coverage in LLM-generated text. Through extensive experimentation using the ClueWeb09 dataset, we demonstrated the framework's effectiveness across different evaluation scenarios, with our best method achieving strong correlation with human judgments. The modular architecture of \framework enables flexible adaptation to various evaluation requirements, from manual to automatic approaches for aspect identifications and alignment. Our results highlighted several key findings: (1) the importance of sophisticated coverage models in improving evaluation accuracy, (2) the viability of automatic evaluation approaches that maintain comparable performance to methods requiring ground truth annotations, and (3) the framework's ability to provide meaningful assessments across different LLM architectures and scales. In future work, individual components of our \framework could be improved in terms of effectiveness and efficiency. Additionally, exploring the relationship between model size, evaluation accuracy, and computational efficiency could provide valuable insights for practical applications. Last but not least, the potential bias introduced by using the same or similar LLM when generating ground-truth aspects should be investigated. By using our metric, other works can explore methods for optimizing LLMs to produce more comprehensive outputs.


\section*{Acknowledgments}
This work was supported in part by the Center for Intelligent Information Retrieval (CIIR), in part by the Office of Naval Research contract number N000142212688, and in part by NSF grants \#2143434 and \#2106282. We acknowledge the support from the Austrian Marshall Plan Foundations, Stefan Wegenkittl, and Martin Uray who made Alexander Krubner's visit to the CIIR possible. Any opinions, findings and conclusions or recommendations expressed in this material are those of the authors and do not necessarily reflect those of the sponsors.

\section*{Limitations}
Our evaluation framework, while showing promising results, suffers from several limitations that should be considered. First, our experiments reveal that even large language models with 70B parameters sometimes struggle with accurate aspect-claim alignment. This suggests that the correlation with human judgments could potentially be improved by specifically optimizing LLMs for this task, either through fine-tuning or more sophisticated prompting strategies.

Second, our current implementation uses zero-shot prompting for query aspect generation without systematic evaluation of this component's effectiveness. Future work should explore methods to optimize and rigorously evaluate the aspect generation process, potentially through human evaluation or comparison with expert-curated aspect sets. This could lead to more reliable and comprehensive aspect coverage assessment.

In addition, there is a potential source of bias when using the same or similar LLM architecture both for generating query aspects and for producing responses for evaluation. This circular dependency might lead to artificially inflated performance metrics if the evaluated model shares similar biases or knowledge patterns with the model used for aspect generation. Future research should investigate the extent of this potential bias and explore methods to mitigate it, such as using diverse model architectures or more comprehensive human-curated aspects for evaluation.

Last but not least, our current evaluation is limited to English-language content using web-based corpora. This narrow focus excludes evaluation of multilingual capabilities and limits the framework's applicability to other languages and cultures. Additionally, the reliance on web corpora may not be suitable for domains requiring specialized knowledge bases or authoritative sources. Future work should explore extending \framework to support multilingual evaluation and integration with domain-specific knowledge bases to broaden its applicability across different languages, cultures, and specialized fields.

\bibliography{references}

\appendix
\section{Appendix}

\subsection{Atomic Claim Generation}
\label{app:claim-gen}
Our atomic claim generation process transforms complex text into atomic, verifiable statements while preserving essential context, as shown in Figure \ref{fig:claim-generator}. The process focuses on creating decontextualized, self-contained claims that each express a single verifiable fact. When breaking down complex sentences, the process maintains important qualifiers, conditions, and temporal information while replacing contextual references with their explicit referents. For instance, a complex sentence about coffee's effects would be decomposed into separate claims about its components and their individual effects, with each claim being fully self-contained and independently verifiable.

\subsection{Model Performance Analysis}
\label{app:model-perf}
The performance analysis presented in Tables \ref{tab:model-performance-cs2-cs3-8b} and \ref{tab:retrieval-models} reveals several important patterns in model behavior. When comparing the 8B and 70B versions of Llama-3.1 as coverage models, we observe that the 8B model generally produces higher raw coverage scores, while the 70B model demonstrates stronger correlation with human judgments. This suggests a trade-off between computational efficiency and evaluation accuracy. The retrieval model comparison shows consistent advantages for dense retrieval methods over traditional BM25, with the Snowflake-Arctic-Embed models showing particular strength in handling queries where simple lexical matching is insufficient. Web-based retrieval consistently produces higher factuality scores compared to corpus-based approaches across all tested models. Table \ref{tab:retrieval-models} analyzes the impact of different retrieval models on the correlation between \framework variants and human judgments. The results show that larger dense retrieval models (Snowflake-Arctic-Embed-L) consistently outperform traditional BM25 across all \framework variants, with improvements particularly notable in \framework 2 and 3.

\begin{table*}[t]
    \footnotesize
    \centering
    \begin{tabular}{lllcccc}
    \toprule
    \multirow{2}{*}{\textbf{LLM}} & \multirow{2}{*}{\textbf{Retrieval}} & \multirow{2}{*}{$S_\text{fact}$} & \multicolumn{2}{c}{\textbf{\framework-S}} & \multicolumn{2}{c}{\textbf{\framework-A}} \\
    \cmidrule(lr){4-5} \cmidrule(lr){6-7}
    & & & $S_\text{coverage}$ & $\text{ICAT-S}_1$ & $S_\text{coverage}$ & $\text{ICAT-A}_1$ \\
    \midrule
    \multirow{2}{*}{GPT-4} 
    & Corpus & 0.343 & 0.563 & 0.394 & 0.563 & 0.394 \\
    & Web Search & 0.756 & 0.583 & 0.658 & 0.583 & 0.658 \\
    \midrule
    \multirow{2}{*}{Llama-3-70B}
    & Corpus & 0.327 & 0.616 & 0.401 & 0.616 & 0.401 \\
    & Web Search & 0.730 & 0.636 & 0.679 & 0.636 & 0.679 \\
    \midrule
    \multirow{2}{*}{Mixtral-8x22B}
    & Corpus & 0.344 & 0.567 & 0.401 & 0.567 & 0.401 \\
    & Web Search & 0.764 & 0.587 & 0.664 & 0.587 & 0.664 \\
    \midrule
    \multirow{2}{*}{Openchat 3.5 (7B)}
    & Corpus & 0.340 & 0.550 & 0.391 & 0.550 & 0.391 \\
    & Web Search & 0.752 & 0.570 & 0.648 & 0.570 & 0.648 \\
    \bottomrule
    \end{tabular}
    \caption{Comprehensive evaluation results comparing ICAT-S and ICAT-A variants using Llama-3.1-8B as the coverage model. Results show performance across different LLMs using both corpus-based and web-based retrieval methods, demonstrating consistent improvements in factuality and coverage scores when using web search compared to corpus-based retrieval. The table highlights how different retrieval methods affect both factuality ($S_\text{fact}$) and coverage scores across various model architectures.}
    \label{tab:model-performance-cs2-cs3-8b}
\end{table*}

\begin{figure*}[h!]
    \centering
    \includegraphics[width=0.9\textwidth]{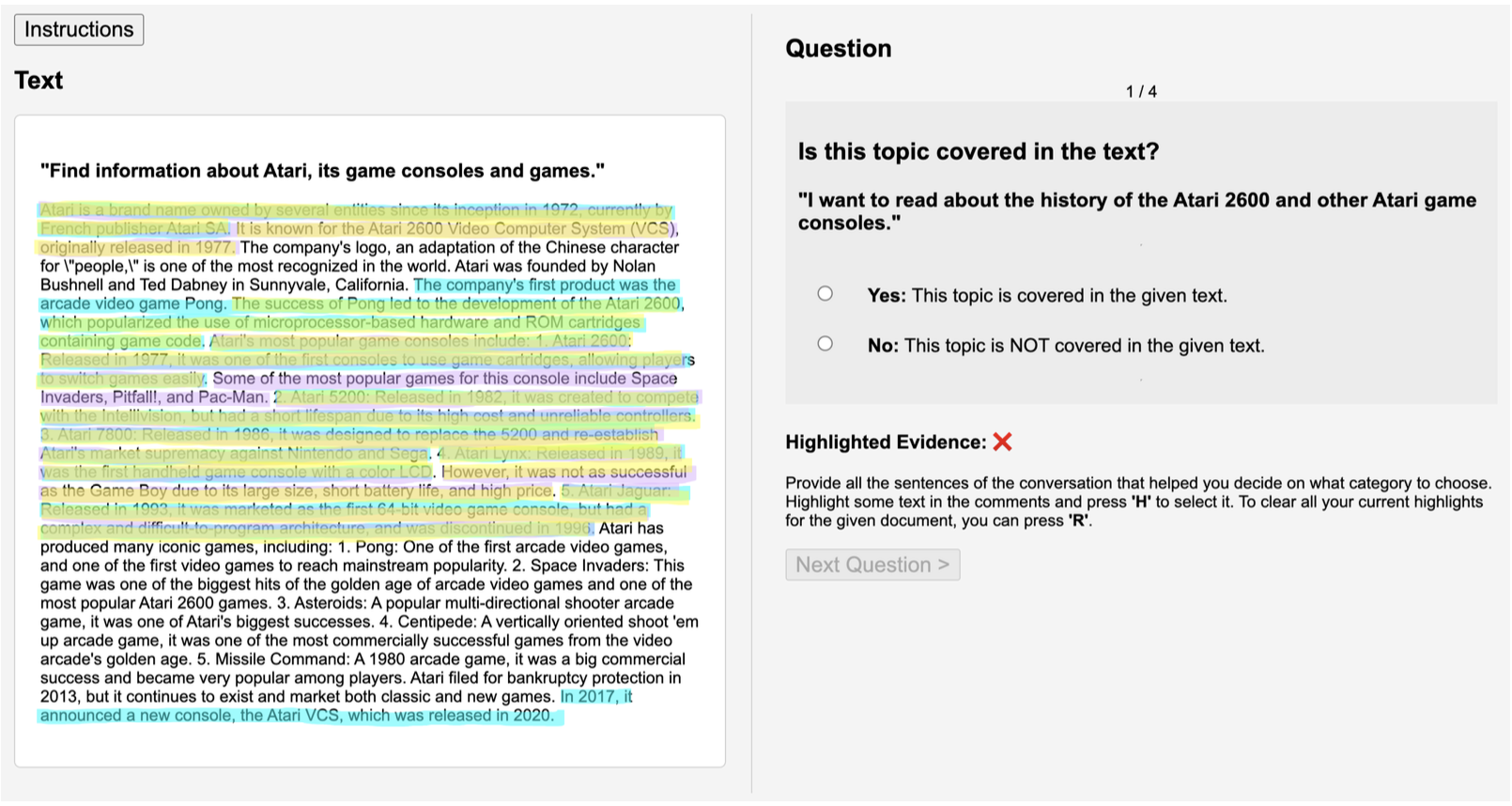}
    \caption{Human annotation interface showing the supporting evidence highlights from 3 annotators for the given query in the shown LLM response.}
    \label{fig:annotation-ui}
\end{figure*}

\subsection{Prompting Details}
\label{app:prompts}
\begin{figure*}[h!]
    \centering
    \includegraphics[width=0.735\textwidth]{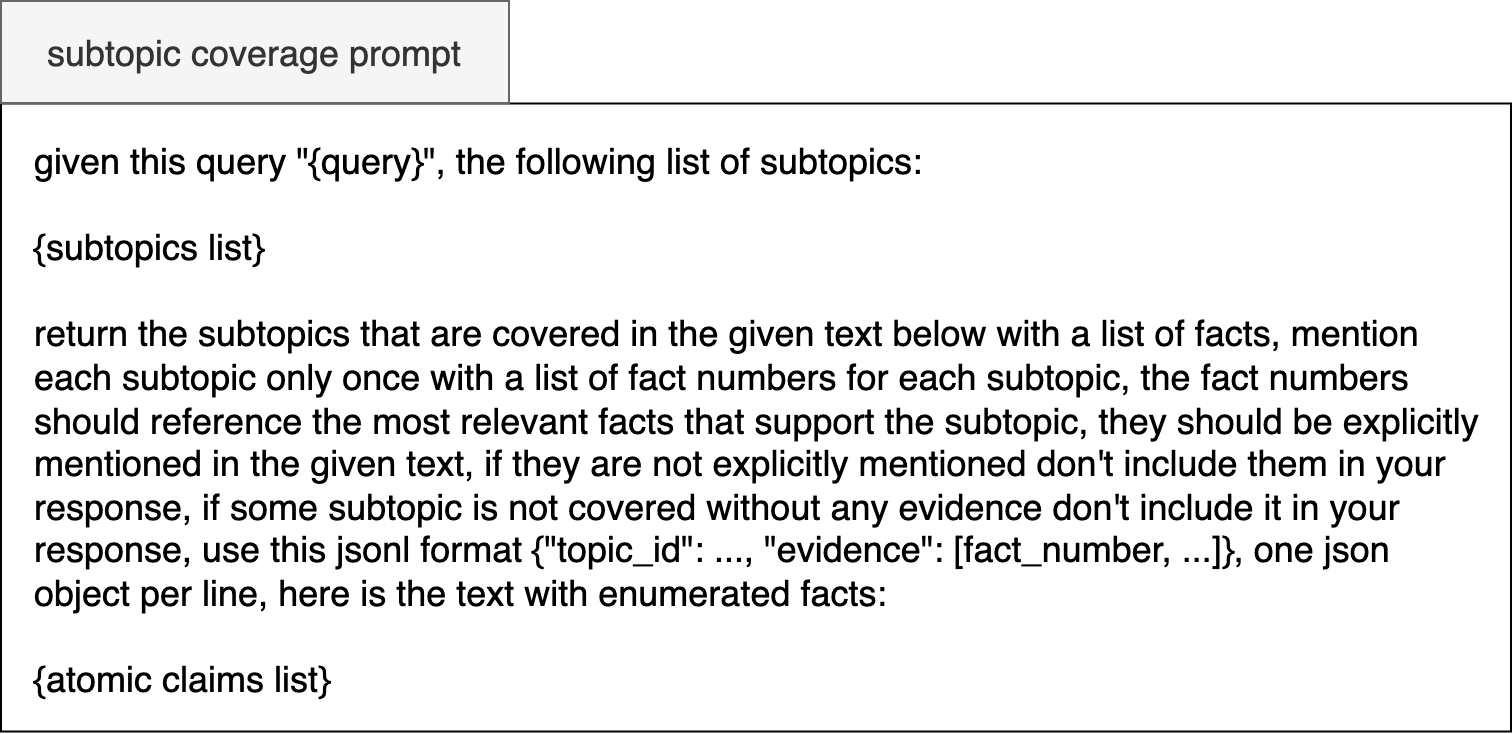}
    \caption{Subtopic coverage prompt used to identify which subtopics are covered in a given text. The prompt instructs the model to analyze a list of atomic claims and return covered subtopics with supporting evidence in a structured JSON format. Each response must include topic IDs and corresponding fact numbers that explicitly appear in the text.}
    \label{fig:coverage-prompt}
\end{figure*}

\begin{figure*}[h!]
    \centering
    \includegraphics[width=0.73\textwidth]{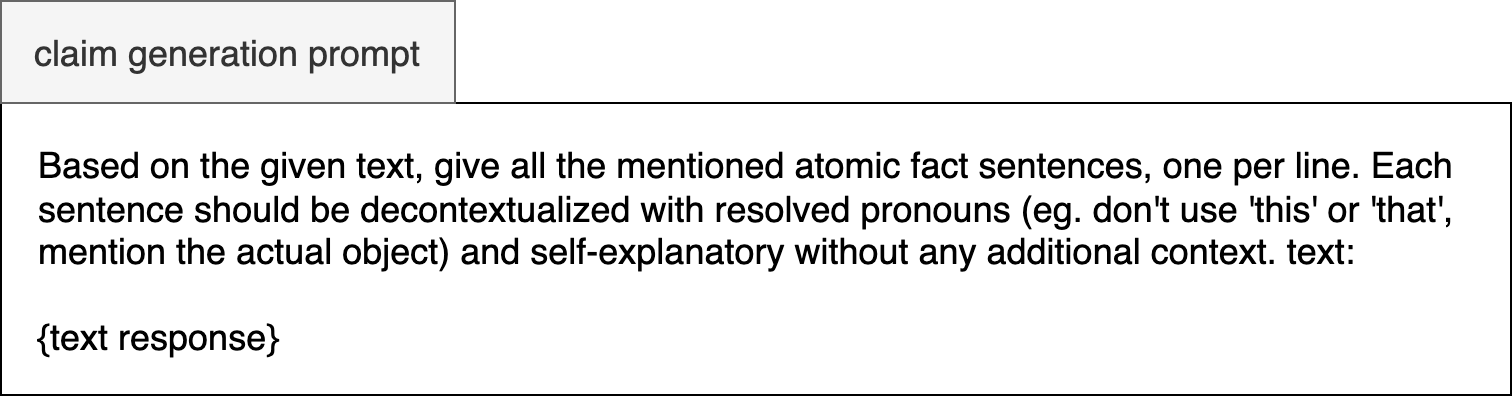}
    \caption{Claim generation prompt designed to extract atomic factual statements from input text. The prompt emphasizes the importance of creating decontextualized, self-explanatory sentences with resolved pronouns and independent context, ensuring each generated claim can stand alone as a verifiable statement.}
    \label{fig:atomic-claim-prompt}
\end{figure*}

\begin{figure*}[h!]
    \centering
    \includegraphics[width=0.73\textwidth]{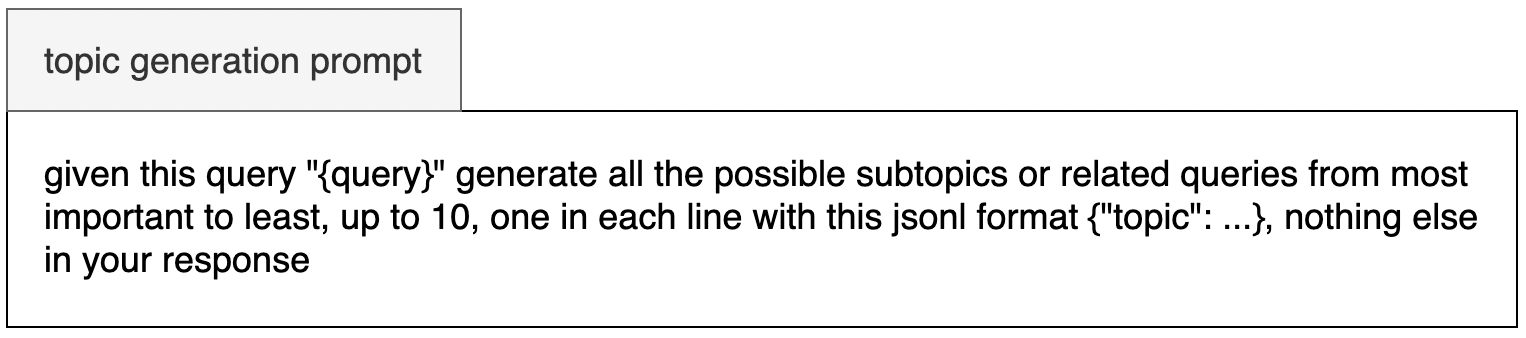}
    \caption{Topic generation prompt used to automatically generate diverse subtopics for a given query. The prompt requests possible subtopics or related queries ordered by importance, with output formatted as JSON objects and limited to a maximum of 10 topics. This automated approach enables reference-free evaluation when ground-truth topics are unavailable.}
    \label{fig:topic-generation-prompt}
\end{figure*}

The framework employs three specialized prompts, each carefully designed for its specific task. The subtopic coverage assessment prompt (Figure \ref{fig:coverage-prompt}) requests the identification of covered subtopics from a given list, requiring evidence in the form of fact numbers that explicitly appear in the text. The prompt specifies a structured JSON output format where each line contains a topic ID and its supporting evidence. The atomic claim generation prompt (Figure \ref{fig:atomic-claim-prompt}) focuses on extracting decontextualized, self-explanatory fact sentences from the input text, emphasizing the importance of resolved pronouns and independent context. The topic generation prompt (Figure \ref{fig:topic-generation-prompt}) elicits possible subtopics or related queries for a given query, requiring them to be ordered by importance and formatted as JSON objects, with a maximum of 10 topics.

\subsection{Human Evaluation Interface}
\label{app:human-annotation}
The human evaluation interface shown in Figure \ref{fig:annotation-ui} was developed to facilitate consistent and accurate topic coverage assessment. The interface presents the query and response in a split-screen layout, enabling annotators to highlight evidence for different topics using a color-coded system. To ensure annotation quality, we implemented minimum time requirements per response and included attention check questions. Annotators were guided to read responses completely before beginning their annotations and to identify text spans that provide direct evidence for topic coverage.

\subsection{Component-level Evaluation}
\label{app:component-eval}

\textbf{Atomic Claim Generation Quality.} We evaluated the quality of atomic claim generation by comparing our fine-tuned Llama-3.1-8B model against the zero-shot Llama-3.1-70B model on 50 randomly sampled queries. Human annotators created ground-truth atomic facts for each query, and we assessed the generated claims using multiple metrics. To enforce local comparisons between atomic facts, we used a sliding window approach (window size ±3) when computing similarity metrics. 

We evaluated using several metrics: (1) \textit{BERTScore}\footnote{using microsoft/deberta-v3-base} metrics for token-level similarity; (2) \textit{Semantic Precision/Recall/F1} computed using cosine similarities of sentence embeddings\footnote{using Alibaba-NLP/gte-base-en-v1.5}, where precision measures how many generated claims have semantically similar ground-truth claims, recall measures coverage of ground-truth claims, and F1 is their harmonic mean; (3) \textit{Completeness}, similar to semantic recall but with similarities binarized at threshold 0.8 to measure hard coverage; (4) traditional metrics (ROUGE, BLEU); and (5) \textit{Fact Precision}, which measures the proportion of atomic claims that are properly formulated—specifically, claims that are decontextualized, self-contained, and verifiable in isolation without pronouns or missing context based on human judgements.

As shown in Table~\ref{tab:claim-generation-eval}, the 70B model achieves higher scores on most semantic similarity metrics (e.g., BERTScore F1: 0.571 vs 0.499), suggesting better overall alignment with ground-truth facts. However, the fine-tuned 8B model demonstrates superior fact precision (0.838 vs 0.753), indicating better decontextualization and self-containment of claims—crucial properties for accurate retrieval-based grounding. This trade-off between semantic similarity and proper claim formulation justifies our choice of using a fine-tuned smaller model for efficient claim generation.

\begin{table}[ht!]
\centering
\scriptsize
\begin{tabular}{l|cc|c}
\toprule
\textbf{Metric} & \textbf{Llama-3.1-8B} & \textbf{Llama-3.1-70B} & \textbf{Diff} \\
\midrule
BERTScore Precision & 0.497 & 0.584 & +0.088 \\
BERTScore Recall & 0.514 & 0.569 & +0.055 \\
BERTScore F1 & 0.499 & 0.571 & +0.072 \\
Semantic Precision & 0.618 & 0.726 & +0.109 \\
Semantic Recall & 0.808 & 0.825 & +0.017 \\
Semantic F1 & 0.681 & 0.761 & +0.080 \\
Completeness & 0.537 & 0.593 & +0.056 \\
ROUGE-1 F & 0.703 & 0.706 & +0.003 \\
ROUGE-2 F & 0.555 & 0.563 & +0.008 \\
ROUGE-L F & 0.599 & 0.610 & +0.011 \\
BLEU & 0.379 & 0.391 & +0.012 \\
Fact Precision & 0.838 & 0.753 & -0.085 \\
\bottomrule
\end{tabular}
\caption{Evaluation of atomic claim generation quality comparing fine-tuned 8B and zero-shot 70B models. The 8B model shows superior fact precision despite lower semantic similarity scores.}
\label{tab:claim-generation-eval}
\end{table}

\textbf{Topic Generation Quality.} We assessed the relevance of automatically generated subtopics for the same 50 queries. Human annotators evaluated each generated subtopic as relevant or irrelevant to the query. Table~\ref{tab:topic-generation-eval} shows that both 8B and 70B models achieve remarkably high subtopic precision (0.966 and 0.958, respectively), demonstrating that even smaller models can effectively generate relevant query aspects when properly prompted. This validates our approach of using LLM-generated topics in \framework-A when ground-truth aspects are unavailable.

\begin{table}[ht!]
\centering
\scriptsize
\begin{tabular}{l|c}
\toprule
\textbf{Model} & \textbf{Subtopic Precision} \\
\midrule
Llama-3.1-8B & 0.966 \\
Llama-3.1-70B & 0.958 \\
\bottomrule
\end{tabular}
\caption{Human evaluation of topic generation quality. Both models achieve high precision in generating relevant subtopics.}
\label{tab:topic-generation-eval}
\end{table}

\textbf{Coverage Assessment Accuracy.} To evaluate the accuracy of our aspect-claim alignment module, we compared the coverage predictions of \framework-S (using Llama-3.1-70B) against human judgments on the set of ground-truth topics. Table~\ref{tab:coverage-accuracy} shows strong agreement metrics, with high precision indicating that the topics identified as covered by our system are indeed covered according to human judgment, and good recall showing that most human-identified covered topics are captured by our system.

\begin{table}[ht!]
\centering
\scriptsize
\begin{tabular}{l|c}
\toprule
\textbf{Coverage Metric} & \textbf{Value} \\
\midrule
Coverage Precision & 0.903 \\
Coverage Recall & 0.798 \\
Coverage F1 & 0.835 \\
\bottomrule
\end{tabular}
\caption{Accuracy of \framework-S topic coverage predictions compared to human judgments using Llama-3.1-70B as the coverage model.}
\label{tab:coverage-accuracy}
\end{table}

\textbf{NLI Model Performance.} Our claim grounding relies on a DeBERTa-based NLI model for efficiency. Table~\ref{tab:nli-performance} shows the model's performance across multiple NLI benchmarks. The strong performance on MNLI (0.903) and FEVER-NLI (0.777) demonstrates effectiveness on standard entailment and fact verification tasks. While performance on adversarial datasets (ANLI) is lower, as expected, the model still maintains reasonable accuracy even on the challenging ANLI-R3 subset (0.495). These results justify using this efficient model over more expensive LLM-based approaches for claim verification.

\begin{table}[ht!]
\centering
\scriptsize
\begin{tabular}{l|c}
\toprule
\textbf{Dataset} & \textbf{Accuracy} \\
\midrule
MNLI & 0.903 \\
FEVER-NLI & 0.777 \\
ANLI-all & 0.579 \\
ANLI-R3 & 0.495 \\
\bottomrule
\end{tabular}
\caption{Performance of the DeBERTa-based NLI model used for claim grounding across multiple benchmarks.}
\label{tab:nli-performance}
\end{table}

\begin{table*}[ht!]
    \centering
    \footnotesize
    \begin{tabular}{llccc}
    \toprule
    \textbf{Method} & \textbf{Retrieval Model} & \textbf{Pearson's $\rho$} & \textbf{Spearman's $\rho$} & \textbf{Kendall's $\tau$} \\
    \midrule
    \multirow{3}{*}{\framework-M} & BM25 & 0.182 & 0.175 & 0.138 \\
    & Snowflake-Arctic-Embed-M & 0.196 & 0.184 & 0.146 \\
    & Snowflake-Arctic-Embed-L & 0.205 & 0.192 & 0.153 \\
    \midrule
    \multirow{3}{*}{\framework-S} & BM25 & 0.472 & 0.443 & 0.368 \\
    & Snowflake-Arctic-Embed-M & 0.489 & 0.452 & 0.376 \\
    & Snowflake-Arctic-Embed-L & 0.503 & 0.461 & 0.384 \\
    \midrule
    \multirow{3}{*}{\framework-A} & BM25 & 0.242 & 0.225 & 0.187 \\
    & Snowflake-Arctic-Embed-M & 0.251 & 0.236 & 0.191 \\
    & Snowflake-Arctic-Embed-L & 0.256 & 0.234 & 0.195 \\
    \bottomrule
    \end{tabular}
    \caption{Comparative analysis of retrieval model impact on ICAT variants' correlation with human judgments, using Llama-3.1-70B for topic-claim alignment. Results demonstrate the superiority of dense retrieval models (Snowflake-Arctic-Embed) over traditional BM25 across all ICAT variants, with the largest improvements seen in ICAT-S and ICAT-A. The analysis includes three correlation metrics (Pearson's, Spearman's, and Kendall's) to provide a comprehensive view of alignment with human assessments across different statistical measures.}
    \label{tab:retrieval-models}
\end{table*}

\end{document}